# Multiple Instance Learning by Discriminative Training of Markov Networks

**Hossein Hajimirsadeghi, Jinling Li, Greg Mori**
School of Computing Science
Simon Fraser University

**Mohamed Zaki, Tarek Sayed**
Department of Civil Engineering
University of British Columbia

## Abstract

We introduce a graphical framework for multiple instance learning (MIL) based on Markov networks. This framework can be used to model the traditional MIL definition as well as more general MIL definitions. Different levels of ambiguity – the portion of positive instances in a bag – can be explored in weakly supervised data. To train these models, we propose a discriminative max-margin learning algorithm leveraging efficient inference for cardinality-based cliques. The efficacy of the proposed framework is evaluated on a variety of data sets. Experimental results verify that encoding or learning the degree of ambiguity can improve classification performance.

## 1 INTRODUCTION

Multiple instance learning aims to recognize patterns from weakly supervised data. Contrary to standard supervised learning, where each training instance is labeled, in the MIL paradigm training instances are given in positive and negative bags. In the traditional MIL definition, a bag is positive if it contains at least one positive instance, while in a negative bag all the instances are negative. This ambiguity in the instance labels is passed on to the learning algorithm, which should incorporate the information to classify unseen bags. In this paper we develop a novel framework for MIL with a more general definition of a positive bag.

Multiple instance learning has been successfully used in many applications such as image categorization (Chen et al., 2006), text categorization (Andrews et al., 2002), content-based image retrieval (Zhang and Goldman, 2002), text-based image retrieval (Li et al., 2011; Duan et al., 2011), object detection (Viola et al., 2006) and tracking (Babenko et al., 2009). Chen et al. (2006) treated each image as a bag of instances corresponding to blocks, regions, or patches of the image for the purpose of image categorization. Andrews et al. (2002) approached text categorization with a MIL framework, where each document is represented by a bag of passages. Li et al. (2011) and Duan et al. (2011) used MIL to handle ambiguity in labels of training images incurred by coarse ranking of web images. Viola et al. (2006) used MIL to overcome the ambiguity in object annotation, by representing each image with a bag of windows centered around the ground truth. Likewise, in object tracking Babenko et al. (2009) used several blocks around the estimated object location to construct a positive training bag for MIL.

The traditional MIL definition states that *at least one* of the instances in a positive bag is positive. However, this is a too weak statement in many MIL applications. For example, in image retrieval most top-ranked training images are truly relevant to the query – they are true positives and not just additional irrelevant elements in a bag (Li et al., 2011). Using this prior information can help to train stronger and more robust classifiers. Further, in some applications, because of noisy, imperfect, or low-quality feature representations, negative bags can contain instances that are effectively indistinguishable from positive instances. In these situations more robust MIL definitions are needed.

To address these issues, we develop a MIL framework based on Markov networks with a flexible notion of a positive bag. This general MIL framework uses cardinality-based measurements over bags, which extend from the notion of "at least one positive" to "at least some positives" to "nearly all positives." Thus, it can explore different levels of ambiguity in the data. In addition, this framework can be adapted to estimate the appropriate MIL notion from training data without prior information about the fraction of positives in the bags. We show that it is possible to use efficient inference techniques (Gupta et al., 2007) to train and evaluate these general MIL models quickly. For the learning criterion, we propose a max-margin discriminative algorithm to train the models.

This paper is organized as follows. Section 2 reviews related work. Section 3 describes our framework of multiple instance learning with Markov networks. In particular, the models for different MIL definitions, including the traditional MIL definition and more general MIL definitions are described in this section. In Section 4 the inference and learning algorithms are explained. Section 5 provides experimental comparisons to state-of-the-art MIL algorithms and an application to video sequence classification. We conclude in Section 6.

## 2 RELATED WORK

Dietterich et al. (1997) introduced the first algorithms for multiple instance learning. The main idea was to construct a hyper-rectangle maximizing the number of enclosed instances from positive bags while excluding all the instances of negative bags. Based on similar ideas, the general diverse density (DD) framework (Maron and Lozano-Pérez, 1998) was proposed. This algorithm works by finding a concept point which is near to at least one instance of every positive bag, but far from all negative instances. Next, EM-DD (Zhang and Goldman, 2002), the expectation-maximization version of DD, was proposed by incorporating the iterative EM approach of estimating positive instances and refining the concept hypothesis within the DD framework.

Gärtner et al. (2002) defined a kernel for multiple instance data and used SVMs to learn a bag classifier. Andrews et al. (2002) modified SVMs for MIL, proposing two algorithms. The first, mi-SVM, aims to maximize the instance margin jointly over the hidden instance labels. The second, MI-SVM, tries to maximize the bag margin, where the bag margin is defined by the most positive instance of each bag. Chen et al. (2006) employed a DD function to map the instances of a bag into a bag-level feature vector. Then, the important features were used by 1-norm SVM for image categorization. Zhou et al. (2009) proposed MIGraph and miGraph. In these methods, first a graph is constructed for each bag, and then an SVM is trained by designing a graph kernel. Thus, by considering the relations among the instances in a bag, the instances are treated as non-i.i.d samples.

The very successful Latent SVM (Felzenszwalb et al., 2010) is also a multiple instance learning framework. For positive instances, a set of latent variable values is used. One can consider the set of completed data instances (latent variable values with observed input feature values) as a "bag" in MIL, as in the MI-SVM framework (Andrews et al., 2002). Latent SVMs have been used in numerous applications, and often obtain state of the art performance. However, they use the "at least one positive instance" positive bag definition. As noted above, for some applications this is limiting since many latent variable settings could in fact be positive and could aid in training a better classifier. The more general MIL definition and algorithms in this paper aim to remedy this.

In recent years, more advanced algorithms have been developed to address non-traditional MIL definitions. Gehler and Chapelle (2007) proposed SVM-like algorithms, AL-SVM and AW-SVM, for MIL. They argued that different levels of ambiguity in positive bags can influence the performance of MIL-based methods. Hence, they provided the possibility to encode prior knowledge about the data set, i.e., fraction of positive instances (witnesses) in a bag. However, these algorithms need a preset parameter which determines the fixed ratio of witnesses.

Bunescu and Mooney (2007) used the transductive SVM framework to propose a MIL algorithm for sparse positive bags. Li and Sminchisescu (2010) proposed a MIL model based on likelihood ratio estimation. The likelihood ratio is estimated by a support vector regression scheme. For bag classification, an SVM is trained to linearly combine the instance likelihood ratios in a postprocessing step. Although, the original model formulation follows the traditional MIL assumption, however, their experiments show that the postprocessing makes this algorithm suboptimally adaptive to different witness rates.

Duan et al. (2011) and Li et al. (2011) introduced a generalized MIL definition, where the positive bags contained at least a certain portion of positive instances. They used a mixed-integer SVM formulation with new constraints on instance labels of the bags. It is shown that this NP-hard problem can be viewed as a multiple kernel learning problem with an exponential number of base kernels. Thus, this algorithm requires some heuristics to solve the original problem. Hajimirsadeghi and Mori (2012) proposed a boosting algorithm for MIL, which can softly explore different levels of ambiguity using linguistic aggregation functions with different degrees of orness. However, this algorithm also needs approximate before-hand knowledge of the witness ratio, or uses cross-validation to estimate it.

In this work, we propose a MIL framework based on Markov networks. This framework is used to model more general MIL definitions, and superior to previous algorithms (Gehler and Chapelle, 2007; Duan et al.,

2011; Li et al., 2011; Hajimirsadeghi and Mori, 2012), it can also work without prior information about the fraction of positive instances inside the bags. In fact, the proposed model can be trained to discover this knowledge directly from data. The inference and learning of the proposed models is exact and no heuristics are needed. Further, the Markov network allows flexible modification and extensions, for instance modeling bag structure. This framework could also be modified to address issues such as training individual classifiers from group statistics of label proportions (Kueck and de Freitas, 2005; Quadrianto et al., 2009; Rueping, 2010).

Note that there are also some other MIL methods based on Markov networks or conditional random fields (CRFs). Deselaers and Ferrari (2010) proposed MI-CRF. In this method, the bags are modelled as nodes in a CRF, where each node can take one of the instances in the bag as its state. So, the bags are jointly trained and classified in this model. Warrell and Torr (2011) proposed another CRF-based method. This method provides a structured bag model, by constructing an undirecetd graph among the instances, instance labels and the bag label. In this CRF, hard and soft MIL constraints are incorporated in the model by defining energy functions between the labels. However, infering the proposed CRFs is performed *approximately* by dual decomposition, and the models are trained by deterministic annealing. Tarlow et al. (2012) proposed a model to approach MIL by CRFs with cardinality potentials over instance labels. However, this model works by sum-product (i.e., marginalization) inference of cardinality potentials. Note that maximum a posteriori (MAP) inference of cardinality potentials, which is used in our proposed method, is faster than the sum-product inference (Tarlow et al., 2012). In addition, our max-margin learning algorithm is different from their maximum likelihood approach to learning.

## 3 MIL USING MARKOV NETWORKS

In MIL, training examples are presented in bags where the instances in a bag share a label. In this work, we use Markov networks to model MIL problems and develop a generalized notion of positive bags.

The Markov network is used to define a scoring function for a bag. A graphical representation of the proposed Markov network for a bag is shown in Figure 1. Each instance and its label are modeled by two nodes in a clique. The clique potential specifies a classifier for an individual instance. A second clique contains all instance labels and the bag label. This clique is used to define what makes a bag positive. Varying this clique potential will lead to different MIL definitions, and is the focus for our work.

### 3.1 MODEL DETAILS

More formally, let $\mathbf{X} = \{\mathbf{x}_1, \cdots, \mathbf{x}_m\}$ denote a bag with $m$ instances and a binary bag label $y \in \{-1, 1\}$. The collective binary instance labels are denoted by $\mathbf{h} = \{h_1, \cdots, h_m\}$. We use the Markov network in Figure 1 to define a scoring function over tuples $(\mathbf{X}, \mathbf{h}, y)$. In testing, this scoring function will be used to find the label $y$ for a test bag, inferring the bag and instance labels that maximize the scoring function.

The network has cliques on each instance and its label, and one clique on all instance labels and the bag label. We define the scoring function on these cliques by:

$$f_\mathbf{w}(\mathbf{X}, \mathbf{h}, y) = \phi^C_\mathbf{w}(\mathbf{h}, y) + \sum_i \phi^I_\mathbf{w}(\mathbf{x}_i, h_i), \quad (1)$$

where $\phi^I_\mathbf{w}(\mathbf{x}_i, h_i)$ represents the potential between each instance and its label, and $\phi^C_\mathbf{w}(\mathbf{h}, y)$ is the clique potential over all the instance labels and the bag label. Note that the potential functions in (1) are parametrized by $\mathbf{w}$. We explain the details of these potential functions as follows.

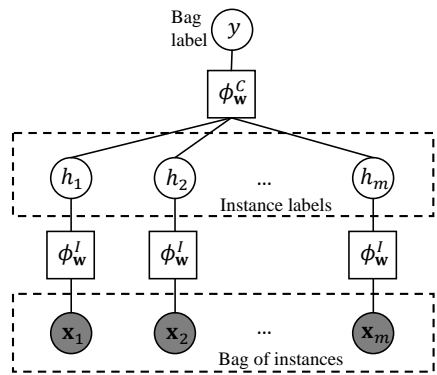

Figure 1: Graphical illustration of the proposed model for multiple instance learning. Potential functions relate instances $\mathbf{x}_i$ to labels $h_i$. A clique relates all instance labels $h_i$ to the bag label $y$.

**Instance-Label Potential** $\phi^I_\mathbf{w}(\mathbf{x}_i, h_i)$: This potential function models the compatibility between the $i$th instance feature vector $\mathbf{x}_i$ and its label $h_i$. It is parametrized as:

$$\begin{aligned}
\phi^I_\mathbf{w}(\mathbf{x}_i, h_i) &= \mathbf{w}_I^\top \mathbf{x}_i \, h_i \\
&= \mathbf{w}_I^\top \psi_I(\mathbf{x}_i, h_i).
\end{aligned}$$

**Labels Clique Potential** $\phi_{\mathbf{w}}^C(\mathbf{h}, y)$: This potential function models the relations between the instance labels and the bag label. Since the MIL problems are defined based on the number of positive and negative instances, we can formulate this as a *cardinality-based* clique potential. Cardinality-based potentials are only a function of counts – in this case, the counts of the numbers of positive and negative instances in the bag.

By modifying the form of the cardinality-based potential, we can obtain different MIL definitions, which will be shown in the subsequent section. Moreover, while for arbitrary clique potentials inference could be NP-complete, for cardinality-based potentials efficient inference algorithms exist. This will lead to efficient algorithms for training and testing, described in Section 4.

In order to define the cardinality-based potentials, we will use the notation $m^+/m^-$ for the number of labels in $\mathbf{h}$ which are positive/negative. The clique potential depends on these counts, and the bag label $y$. We parameterize two different clique potentials, one for positive bags ($C_{\mathbf{w}}^+$) and one for negative bags ($C_{\mathbf{w}}^-$):

$$\begin{aligned}\phi_{\mathbf{w}}^C(\mathbf{h}, y) &= C_{\mathbf{w}}\left(m^+, m^-, y\right) \\ &= C_{\mathbf{w}}^+\left(m^+, m^-\right)\mathbb{1}(y=1) \\ &\quad + C_{\mathbf{w}}^-\left(m^+, m^-\right)\mathbb{1}(y=-1).\end{aligned} \qquad (2)$$

The following sections define functions $C_{\mathbf{w}}^+$ and $C_{\mathbf{w}}^-$ that lead to a variety of MIL models.

### 3.1.1 Multiple Instance Markov Network (MIMN)

This network models the standard MIL problem, i.e., in a positive bag at least one of the instances is positive, and in a negative bag all the instances are negative. The labels clique potential is given by

$$C_{\mathbf{w}}^+(0, m) = -\infty \qquad (3)$$
$$C_{\mathbf{w}}^+(m^+, m - m^+) = w_c^+ \quad m^+ = 1, \cdots, m \qquad (4)$$
$$C_{\mathbf{w}}^-(0, m) = w_c^- \qquad (5)$$
$$C_{\mathbf{w}}^-(m^+, m - m^+) = -\infty \quad m^+ = 1, \cdots, m. \qquad (6)$$

This clique potential states that in a positive bag it is impossible to have all the instances be negative (3), but there is the same potential of having more than one positive instance (4). However, for a negative bag, it is only possible to have negative instances (5) & (6). One might set $w_c^+$ and $w_c^-$ to a constant value (e.g. 0), but we treat them as the model parameters and show how to learn them in Section 4.2.

### 3.1.2 Ratio-constrained Multiple Instance Markov Network (RMIMN)

Ratio-constrained MIL extends the notion of positive bags in MIL. In RMIMN, each positive bag contains at least a certain portion of positive instances. For example, at least $x\%$ of the instances should be positive in a positive bag. To model this problem with our proposed Markov network, we can refine the functions $C^+$ and $C^-$:

$$\begin{aligned}C_{\mathbf{w}}^+(m^+, m - m^+) &= -\infty & 0 \leq \frac{m^+}{m} < \rho \\ C_{\mathbf{w}}^+(m^+, m - m^+) &= w_c^+ & \rho \leq \frac{m^+}{m} \leq 1 \\ C_{\mathbf{w}}^-(m^+, m - m^+) &= w_c^- & 0 \leq \frac{m^+}{m} < \rho \\ C_{\mathbf{w}}^-(m^+, m - m^+) &= -\infty & \rho \leq \frac{m^+}{m} \leq 1,\end{aligned} \qquad (7)$$

where $\rho$ indicates the required portion of positive instances in a positive bag.

### 3.1.3 Generalized Multiple Instance Markov Network (GMIMN)

GMIMN allows a very flexible notion of positive bags. We allow the portions of positive and negative instances in bags to be a learned parameter, discovered from the data. The MIL model will learn which fractions of instances tend to be positive in a bag. This network provides a very general model for multiple instance learning and is parametrized by:

$$\begin{aligned}C_{\mathbf{w}}^+(0, m) &= -\infty \\ C_{\mathbf{w}}^+(m^+, m - m^+) &= \sum_{k=1}^{K} w_k^+ \mathbb{1}\left(\frac{k-1}{K} < \frac{m^+}{m} \leq \frac{k}{K}\right) \\ & \quad m^+ = 1, \cdots, m \\ C_{\mathbf{w}}^-(m^+, m - m^+) &= \sum_{k=1}^{K} w_k^- \mathbb{1}\left(\frac{k-1}{K} \leq \frac{m^+}{m} < \frac{k}{K}\right) \\ & \quad m^+ = 0, \cdots, m-1 \\ C_{\mathbf{w}}^-(m, 0) &= -\infty.\end{aligned} \qquad (8)$$

where $K$ determines the number of weighted segments of a bag. This model divides the bag size into $K$ equal parts, and the weight of each segment $w_k$ determines how important it is that the number of positive instances be placed inside that interval[1]. In other words, these learning weights specify the importance or impact of different witness ratios for labeling a bag as positive or negative. Large values of $K$ provide more detailed and specific models of bag definition by learning cardinality-based measures with finer resolution, while low values of $K$ define a coarser model of bag. So, by controlling the granularity, this parameter is set in a trade-off between training accuracy and generalization ability[2]. Note that $C_{\mathbf{w}}^+(0, m) = -\infty$ and $C_{\mathbf{w}}^-(m, 0) = -\infty$ are the only required prior information in this model.

---
[1] Note that the weights $w_k$ are not necessarily monotonically increasing as $k$ increases. For example, in a MIL data set, there might be only a few true positive instances in the positive bags, and so the potential of having many instances be positive is low.

With these definitions, we note that using $C_{\mathbf{w}}^+$ and $C_{\mathbf{w}}^-$ defined in any of the MIMN, RMIMN, or GMIMN models makes the clique potential (i.e., $\phi_{\mathbf{w}}^C$) a linear function of the learning parameters. More formally:

$$\phi_{\mathbf{w}}^C(\mathbf{h}, y) = \mathbf{w}_C^\top \Psi_C(\mathbf{h}, y) + g_C(\mathbf{h}, y), \qquad (9)$$

where $\mathbf{w}_C$ represents the concatenation of the learning parameters in $C_{\mathbf{w}}^+$ and $C_{\mathbf{w}}^-$, while $\Psi_C(\mathbf{h}, y)$ and $g_C(\mathbf{h}, y)$ are functions independent of $\mathbf{w}$, which are specified by aggregation of the indicator functions.

## 4 INFERENCE AND LEARNING

The MIL models above define scoring functions that consider counts of instance labels in a bag. Using this, for a given bag we can define a scoring function for labeling a bag $\mathbf{X}$ with a label $y$:

$$F_{\mathbf{w}}(\mathbf{X}, y) = \max_{\mathbf{h}} f_{\mathbf{w}}(\mathbf{X}, \mathbf{h}, y). \qquad (10)$$

Below, we describe how to use efficient inference algorithms (Gupta et al., 2007) to efficiently solve this inference problem for the cardinality-based cliques we defined above.

Using this inference technique, learning can be performed using a max-margin criterion, as in the Latent SVM approach.

Classification of a new test bag can be done in a similar manner. We can predict the bag label by simply running inference, trying $y = +1$ and $y = -1$ and taking the maximum scoring bag label:

$$y^\star = \arg\max_y F_{\mathbf{w}}(\mathbf{X}, y). \qquad (11)$$

### 4.1 INFERENCE

The inference problem is to find the best set of instance labels $\mathbf{h}$ given observed values for the data instances $\mathbf{X}$ and the bag label $y$ – the maximization problem in

---

[2] In the experiments of this paper, we use cross-validation on the values $K = 3$, $K = 5$, and $K = 10$ to roughly estimate this parameter.

(10). Using (1) and (2), the inference problem can be written as

$$\max_{\mathbf{h}} \sum_i \phi_{\mathbf{w}}^I(\mathbf{x}_i, h_i) + C_{\mathbf{w}}(m^+, m^-, y). \qquad (12)$$

This problem is an instance of inference in graphical models with cardinality-based clique potentials (Gupta et al., 2007). This class of clique potentials is specified by two parts: the sum of individual node potentials and a function over all the nodes which only depends on the counts of the nodes which get specific labels. Efficient inference algorithms have been proposed for this class of graphical model in (Gupta et al., 2007). In this paper, we only work with the binary case (i.e., $h_i \in \{+1, -1\}$), for which there is an exact inference algorithm with $O(m \log m)$ time complexity. The inference algorithm is as follows.

First, sort the instances in decreasing order of $\phi_{\mathbf{w}}^I(\mathbf{x}_i, +1) - \phi_{\mathbf{w}}^I(\mathbf{x}_i, -1)$. Then, for $k = 0, \cdots, m$, compute $s_k$, the sum of the top-$k$ instance potentials $\phi_{\mathbf{w}}^I(\mathbf{x}_i, +1) - \phi_{\mathbf{w}}^I(\mathbf{x}_i, -1)$ plus the clique potential $C_{\mathbf{w}}(k, m-k, y)$. Finally, find $k^\star$ which gets the largest $s_k$, and inference is accomplished by assigning the top $k^\star$ instances to positive labels and the rest to negative labels.

### 4.2 LEARNING

The training set is given by $\{(\mathbf{X}^1, y^1), \cdots, (\mathbf{X}^N, y^N)\}$, and the goal is to train the Markov models by learning the parameters $\mathbf{w}$. Inspired by the relations to latent SVM, we formulate the learning problem as minimizing the regularized hinge loss function:

$$\begin{aligned}
&\min_{\mathbf{w}} \sum_{n=1}^N (\mathcal{L}^n - \mathcal{R}^n) + \frac{\lambda}{2} \|\mathbf{w}\|^2 \\
&\text{where } \mathcal{L}^n = \max_y \max_{\mathbf{h}} \left( \Delta(y, y^n) + f_{\mathbf{w}}(\mathbf{X}^n, \mathbf{h}, y) \right), \\
&\mathcal{R}^n = \max_{\mathbf{h}} f_{\mathbf{w}}(\mathbf{X}^n, \mathbf{h}, y^n), \\
&\Delta(y, y^n) = \begin{cases} 1 & \text{if } y \neq y^n \\ 0 & \text{if } y = y^n. \end{cases}
\end{aligned} \qquad (13)$$

One approach to solve this problem approximately is the iterative algorithm of alternating between inference of the latent variables and optimization of the model parameters. So, the first step estimates the instance labels and the second step learns a standard SVM classifier given the estimated instance labels. It can be shown that using this approach with the MIMN model leads to an algorithm very similar to mi-SVM (Andrews et al., 2002).

However, we use the non-convex cutting plane method (Do and Artières, 2009) to directly solve the optimization problem in (13). This method is proved to converge to a local optimum, unlike the heuristic iterative algorithm of mi-SVM, which has no convergence guarantee. The non-convex cutting plane method iteratively makes an increasingly accurate piecewise quadratic approximation of the objective function. At each iteration, a new linear cutting plane is obtained via the subgradient of the objective function and added to the piecewise quadratic approximation. To use this algorithm, the principal issue is to compute the subgradients $\partial_\mathbf{w} \mathcal{L}^n(\mathbf{w})$ and $\partial_\mathbf{w} \mathcal{R}^n(\mathbf{w})$. To this end, we need to know the subgradient of the network potential function, i.e., $\partial_\mathbf{w} f_\mathbf{w}(\mathbf{X}, \mathbf{h}, y)$.

It is simple to show that

$$\partial_\mathbf{w} f_\mathbf{w}(\mathbf{X}, \mathbf{h}, y) = \Psi(\mathbf{X}, \mathbf{h}, y), \quad (14)$$

where $\Psi(\mathbf{X}, \mathbf{h}, y) = \left[ \sum_i \psi_I(\mathbf{x}_i, h_i)^\top, \Psi_C(\mathbf{h}, y)^\top \right]^\top$. Using equations (13) and (14), it can be shown that $\partial_\mathbf{w} \mathcal{L}^n(\mathbf{w}) = \Psi(\mathbf{X}^n, \mathbf{h}^\star, y^\star)$, where $(\mathbf{h}^\star, y^\star)$ is the solution to the inference problem:

$$\max_y \max_\mathbf{h} \left( \Delta(y, y^n) + f_\mathbf{w}(\mathbf{X}^n, \mathbf{h}, y) \right). \quad (15)$$

This inference problem can be solved using the algorithm in 4.1. In summary, for $y = 1$ and $y = -1$ we find $\mathbf{h}$ by doing inference on the resulting graphical model (which has cardinality-based clique potential). Then, the $y$ with the highest value gives the predicted bag label $y^\star$.

In the same way, it can be shown that $\partial_\mathbf{w} \mathcal{R}^n(\mathbf{w}) = \Psi(\mathbf{X}^n, \mathbf{h}^\star, y^n)$, where $\mathbf{h}^\star$ is the solution to the inference problem:

$$\max_\mathbf{h} f_\mathbf{w}(\mathbf{X}^n, \mathbf{h}, y^n). \quad (16)$$

## 5 EXPERIMENTS

In this section we describe the evaluation of our MIL models. First, the proposed models are evaluated on MIL benchmark data sets to demonstrate they can achieve state of the art performance on standard datasets. Next, we evaluate the models on a challenging cyclist helmet recognition dataset, and show that flexibility in the portion of positives in a bag can lead to improved classification accuracy.

### 5.1 BENCHMARK DATA SETS

We evaluate the MIL models on five well-known MIL datasets. These benchmark data sets are the *Elephant*, *Fox*, *Tiger* image data sets (Andrews et al., 2002) and *Musk1* and *Musk2* drug activity prediction data sets (Dieterich et al., 1997). In the image data sets, each bag represents an image, and the instances inside the bag represent 230-D feature vectors of different segmented blobs of the image. These data sets contain 100 positive and 100 negative bags. In the MUSK data sets, each bag describes a molecule, and the instances inside the bag represent 166-D feature vectors of the low-energy configurations of the molecule. Musk1 has 47 positive bags and 45 negative bags with about 5 instances per bag. Musk2 has 39 positive bags and 63 negative bags with variable number of instances in a bag, ranging from 1 to 1044 (average 64 instances per bag). Note that in all experiments of this section, we have used normalized data sets, which are obtained by scaling the features of the original data sets[3] to the range $[0, 1]$.

The 10-fold averaged classification accuracies for the MIMN model on different data sets are shown in Table 1. At each trial, we run the non-convex cutting plane algorithm with all the learning weights initialized to 0, roughly optimized $\lambda$, and at most 300 iterations. This table also includes the classification results with different kernel feature maps. For these data sets (especially Musk1 and Musk2), non-linear kernels are commonly used for SVM-like algorithms. For example, in (Andrews et al., 2002) mi-SVM and MI-SVM are trained on Musk1 and Musk2 data sets by RBF kernels. Or in (Ray and Craven, 2005) and (Bunescu and Mooney, 2007) quadratic kernels have shown successful classification results. Since our algorithm works with linear kernels, we exploit the idea of kernel feature maps. We investigate the performance of quadratic features in addition to the feature maps proposed in (Vedaldi and Zisserman, 2012) for homogeneous kernels: intersection, $\chi^2$, and Jensen-Shannon.

Table 1: MIMN classification accuracy with different kernel functions. The best results are marked in bold face.

| Method | Elephant | Fox | Tiger | Musk1 | Musk2 |
|---|---|---|---|---|---|
| MIMN$_{\text{Linear}}$ | 85.5 | 62.5 | **87.0** | 78.3 | 77.6 |
| MIMN$_{\text{Quadratic}}$ | 82.50 | **64.0** | **87.0** | 85.9 | 81.9 |
| MIMN$_{\text{Intersection}}$ | **89.0** | 59.0 | 85.5 | **86.1** | 89.5 |
| MIMN$_{\chi^2}$ | 87.0 | 60.0 | 84.0 | 84.1 | **90.3** |
| MIMN$_{\text{Jensen-Shannon}}$ | 86.0 | 59.0 | 84.5 | 83.7 | 87.4 |

Now, we compare the best of MIMN with state-of-the-art MIL methods in Table 2[4]. The performance of the methods varies depending on the data set. However, MIMN is always among the best methods. More specifically, it achieves the best accuracy in the Elephant, Fox, Tiger and Musk2 data sets.

---

[3]The original data sets are available online at http://www.cs.columbia.edu/~andrews/mil/datasets.html.

Note that the competing methods miGraph and MI-Graph (Zhou et al., 2009) treat the instances as non-i.i.d samples and model correlations among the bags – this incorporates different information into the model, which is not directly present in our approach.

Next, the results of the experiments with the RMIMN model are presented in Table 3. It can be observed that for the image data sets RMIMN cannot improve MIMN significantly. However, for Musk1 and Musk2 substantial performance gains can be made. The reason might be that in an image usually one of the segments is the true segment (positive instance). So, the prior information, at least one of the instances is positive, is likely sufficient. However, in the Musk data sets, more than one configuration of a molecule might be positive. In fact, it has been previously reported (Gehler and Chapelle, 2007; Hajimirsadeghi and Mori, 2012) that the Musk data sets contain many positive instances in each positive bag. This experiment shows that our graphical approach to MIL allows for exploring different levels of ambiguity in the bags in order to enhance classification accuracy.

Table 3: RMIMN classification accuracy with different $\rho$ values, compared with MIMN. All results are based on linear kernel functions.

| Method | $\rho$ | Elephant | Fox | Tiger | Musk1 | Musk2 |
|---|---|---|---|---|---|---|
| MIMN | - | 85.5 | 62.5 | 87.0 | 78.3 | 77.6 |
| RMIMN | 0.1 | 85.5 | 62.0 | 85.5 | 80.4 | 79.9 |
| RMIMN | 0.2 | 83.5 | 61.0 | 85.0 | 88.1 | 82.8 |
| RMIMN | 0.3 | 84.0 | 56.5 | 83.5 | 83.9 | 88.6 |
| RMIMN | 0.4 | 83.5 | 60.0 | 83.0 | 82.7 | 84.6 |
| RMIMN | 0.5 | 83.5 | 59.5 | 83.5 | 86.0 | 86.6 |
| RMIMN | 0.6 | 84.0 | 59.5 | 84.0 | 85.8 | 86.3 |
| RMIMN | 0.7 | 84.5 | 58.0 | 84.0 | 85.0 | 84.5 |
| RMIMN | 0.8 | 84.0 | 57.5 | 83.5 | 83.8 | 82.6 |
| RMIMN | 0.9 | 85.0 | 61.0 | 83.5 | 83.8 | 82.8 |
| RMIMN | 1.0 | 87.5 | 62.5 | 84.5 | 89.1 | 84.8 |

Finally, the results of the experiments with the GMIMN model are provided in Table 4. We evaluate the performance of this model with $K = 10$ weighted segments. It can be observed that although GMIMN gets very weak prior information on the notion of positive bags, by learning the levels of ambiguity in data it outperforms MIMN in most cases.

---

[4]Note that the reported results for some other methods (e.g. mi-SVM and MI-SVM) on different data sets are also based on the most successful kernels.

Table 4: GMIMN classification accuracy, compared with MIMN. All results are based on linear kernel functions.

| Method | Elephant | Fox | Tiger | Musk1 | Musk2 |
|---|---|---|---|---|---|
| MIMN | 85.5 | 62.5 | 87.0 | 78.3 | 77.6 |
| GMIMN($K = 10$) | 89.0 | 61.5 | 86.5 | 87.1 | 81.4 |

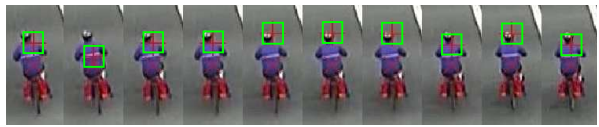

Figure 2: Cyclist helmet classification – is she wearing helmet? how many positives are in this bag? An automatic cyclist detector/tracker is run, with head position estimate in green rectangle. Data instances are features defined on the head position estimates, bags aggregate these over a track.

### 5.2 CYCLIST HELMET RECOGNITION

The previous experiments show that the proposed method is comparable to the state-of-the-art on standard datasets. However, those datasets exhibit limited ambiguity in positive bags. We now show that for more complex situations, our framework can effectively discover the ambiguity in positive bags. In this section, we use our proposed models to address a video classification task. This problem is illustrated in Figure 2. Given an automatically-obtained cyclist trajectory, we must determine whether the cyclist is wearing a helmet or not. One can treat this as a MIL problem – each frame is an instance, and the trajectory forms a bag. The bag (trajectory) should be classified as containing a helmet-wearing cyclist or not. However, the standard MIL or traditional supervised learning approaches (e.g. classify each instance and majority vote) cannot easily handle this problem. Because of imperfection in tracking, it is unlikely that all the instances in a positive bag are truly positive – some will not be well centered on the cyclist's head due to jitter, regardless of the tracker used. Traditional supervised learning would have many corrupted positive instances of helmet-wearing cyclists. Standard MIL would not make full use of the training data, since each track would very likely have more than one positive instance.

#### 5.2.1 Experimental Setup

We work with cyclist trajectories automatically extracted from video data. The data are collected for a busy 4-legged intersection with vehicles, pedestrians, and cyclists, over a two-day period. Kanade-Lucas-Tomasi feature tracking and trajectory clustering are used to extract moving objects. These clusters are then automatically classified (vehicle, pedestrian, cyclist) by analyzing speed profiles (e.g. the pedalling cadence).

Table 2: Comparison between state-of-the-art MIL methods. The best and second best results are highlighted in bold and italic face respectively.

| Method | Elephant | Fox | Tiger | Musk1 | Musk2 |
|---|---|---|---|---|---|
| MIMN | **89** | **64** | **87** | 86 | **90** |
| mi-SVM (Andrews et al., 2002) | 82 | 58 | 79 | 87 | 84 |
| MI-SVM (Andrews et al., 2002) | 81 | 59 | 84 | 78 | 84 |
| MI-Kernel (Gärtner et al., 2002) | 84 | 60 | 84 | 88 | *89* |
| MIRealBoost (Hajimirsadeghi and Mori, 2012) | 83 | *63* | 73 | **91** | 77 |
| MIForest (Leistner et al., 2010) | 84 | **64** | 82 | 85 | 82 |
| MILES (Chen et al., 2006) | 81 | 62 | 80 | 88 | 83 |
| AW-SVM (Gehler and Chapelle, 2007) | 82 | **64** | 83 | 86 | 84 |
| AL-SVM (Gehler and Chapelle, 2007) | 79 | *63* | 78 | 86 | 83 |
| EM-DD (Zhang and Goldman, 2002) | 78 | 56 | 72 | 85 | 85 |
| SVR-SVM (Li and Sminchisescu, 2010) | 85 | 80 | *63* | 88 | 85 |
| MIGraph (Zhou et al., 2009) | 85 | 61 | 82 | *90* | **90** |
| miGraph (Zhou et al., 2009) | *87* | 62 | *86* | *90* | **90** |

We chose a dataset of 24 cyclist tracks for our experiments – 12 wearing helmets and 12 not. The head location is estimated using background subtraction upon the tracks. Samples of tracking the cyclists' heads in the videos are shown in Figure 3. We describe each frame of a track using texton histograms (Malik et al., 2001) in a region of size $20 \times 20$ around the head position (chosen after empirically examining other features). We report the results of helmet classification using leave-one-out cross-validation on this dataset.

We introduce a MIL approach to classify sequences. Each video is treated as a bag of frames represented by instances, and we use the proposed models in Section 3 to classify the bags. We also compare this approach with non-MIL methods. In the non-MIL approach, all frames from positive and negative training videos are put together and labelled according to their video labels. Next, a standard SVM classifier (Chang and Lin, 2011) is trained and used to predict each frame label of the test videos. Finally, the bag label is predicted by one of the following criteria:

- SVM-AtLeastOne: The bag label is positive if at least one of the instance labels is positive.

- SVM-Majority: The bag label is specified by the majority voting of the instance labels.

#### 5.2.2 Experimental Results

The average classification accuracy of each method is shown in Table 5. We include mi-SVM as an additional baseline. The results of the RMIMN model have been provided with different $\rho$ values.

Table 5: Results of the experiments on cyclist helmet classification problem.

| Method | Accuracy % |
|---|---|
| SVM-AtLeastOne | 58.33 |
| SVM-Majority | 79.17 |
| mi-SVM | 62.50 |
| MIMN | 58.33 |
| GMIMN ($K = 5$) | 87.50 |
| RMIMN ($\rho = 0.1$) | 79.17 |
| RMIMN ($\rho = 0.2$) | 83.33 |
| RMIMN ($\rho = 0.3$) | 91.67 |
| RMIMN ($\rho = 0.4$) | 87.50 |
| RMIMN ($\rho = 0.5$) | 91.67 |
| RMIMN ($\rho = 0.6$) | 91.67 |
| RMIMN ($\rho = 0.7$) | 87.50 |
| RMIMN ($\rho = 0.8$) | 87.50 |
| RMIMN ($\rho = 0.9$) | 83.33 |
| RMIMN ($\rho = 1.0$) | 66.67 |

It can be observed that the classification accuracy of SVM-AtLeastOne, MIMN, and mi-SVM are quite low. This shows that the traditional classification approach and MIL definition (used in SVM-AtLeastOne, MIMN, and mi-SVM) are very inefficient in this problem. The traditional MIL definition (i.e., at least one instance of a positive bag is positive) fails because it is very likely that at least one of the instances in a negative bag is classified as positive, and consequently most of the negative bags are assigned positive labels. This problem is due to the imperfection in the classifier and low-quality visual representation of the cyclist's head in the video. However, it is clearly evident that SVM-Majority, RMIMN (with most $\rho$ values), and GMIMN are more robust to these defects. The results show that RMIMN with $\rho = 0.3$, 0.5, and 0.6 outperform all the other methods. Also, it is shown that GMIMN has good performance, learning the MIL definition properly without any prior knowledge of ambiguity level (e.g., parameter $\rho$) and classifying the videos successfully.

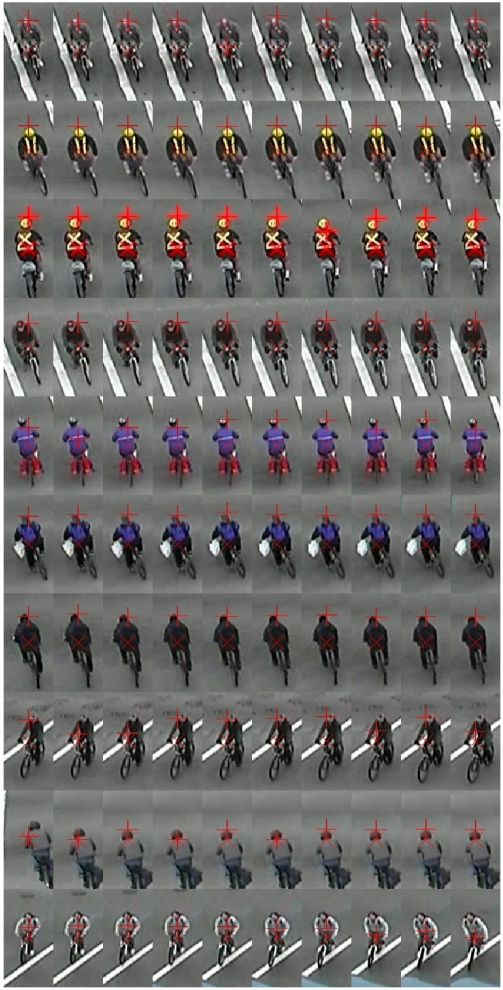

Figure 3: Samples of tracking the cyclists' heads in the videos. Red + shows automatic head position estimate.

## 6 CONCLUSION

We proposed a novel graphical framework for MIL based on Markov networks and max-margin discriminative training. This framework is flexible and can model the traditional MIL definition as well as more general MIL definitions. Thus, it is more robust to the amount of ambiguity (i.e. true positive instances) in the bags. Especially, it can be helpful in vision applications which exhibit imperfect annotation or ambiguous feature representations. For training the proposed models, we formulated the learning process as a max-margin optimization problem.

Experiments on MIL benchmark data sets showed that the proposed algorithm is comparable with state-of-the-art MIL methods. In addition, it was verified that learning and encoding the degree of ambiguity in the classifier can influence the accuracy of classification. We used the proposed framework for classifying cyclist trajectories. This is a challenging problem, where the traditional supervised learning and traditional MIL definitions fail. However, the RMIMN and GMIMN models enhance classification performance by finding more general and robust MIL definitions and mining the degree of ambiguity.

The proposed graphical framework is flexible and can be easily extended or modified. For example, it can be modified to define a bag margin based on the most positive instance of the bag, e.g. MI-SVM (Andrews et al., 2002). It can be also extended for multi-class classification. In addition, more potential functions can be defined between the network nodes. For example, a potential function can be added between a bag-level feature vector and the bag label, or new potential functions can be defined over neighbouring instance labels to treat the instances as non-i.i.d. samples. Finally, this framework could be adapted for individual classification from group statistics, and applied to tasks such as privacy-preserving data mining, election results analysis, spam and fraud detection (Rueping, 2010).

### Acknowledgements

This work was supported by grants from the Natural Sciences and Engineering Research Council of Canada (NSERC) and BCFRST NRAS Research Team Program.

### References

S. Andrews, I. Tsochantaridis, and T. Hofmann. Support vector machines for multiple-instance learning. In *NIPS*, 2002.

B. Babenko, M.H. Yang, and S. Belongie. Visual tracking with online multiple instance learning. In *Computer Vision and Pattern Recognition (CVPR)*, 2009.

R.C. Bunescu and R.J. Mooney. Multiple instance learning for sparse positive bags. In *Proceedings of the 24th International Conference on Machine Learning (ICML)*, pages 105–112. ACM, 2007.

C.C. Chang and C.J. Lin. Libsvm: a library for support vector machines. *ACM Transactions on Intelligent Systems and Technology (TIST)*, 2(3):27, 2011.

Y. Chen, J. Bi, and J.Z. Wang. Miles: Multiple-instance learning via embedded instance selection. *T-PAMI*, 28(12):1931–1947, 2006.

Thomas Deselaers and Vittorio Ferrari. A conditional random field for multiple-instance learning. 2010.

T.G. Dietterich, R.H. Lathrop, and T. Lozano-Pérez. Solving the multiple instance problem with axis-parallel rectangles. *Artificial Intelligence*, 89(1-2):31–71, 1997.

T.M.T. Do and T. Artières. Large margin training for hidden markov models with partially observed states. In *Proceedings of the 26th Annual International Conference on Machine Learning (ICML)*, pages 265–272. ACM, 2009.

L. Duan, W. Li, I. Tsang, and D. Xu. Improving web image search by bag-based re-ranking. *IEEE Transactions on Image Processing*, 20(11):3280–3290, 2011.

P. Felzenszwalb, R. Girshick, D. McAllester, and D. Ramanan. Object detection with discriminatively trained part based models. *PAMI*, 32(9):1627–1645, 2010.

T. Gärtner, P.A. Flach, A. Kowalczyk, and A.J. Smola. Multi-instance kernels. In *Proceedings of the 19th International Conference on Machine Learning (ICML)*, pages 179–186, 2002.

P. Gehler and O. Chapelle. Deterministic annealing for multiple-instance learning. In *AISTATS*, 2007.

R. Gupta, A.A. Diwan, and S. Sarawagi. Efficient inference with cardinality-based clique potentials. In *Proceedings of the 24th International Conference on Machine Learning (ICML)*, pages 329–336. ACM, 2007.

H. Hajimirsadeghi and G. Mori. Multiple instance real boosting with aggregation functions. In *Proceedings of the 21st International Conference on Pattern Recognition*, 2012.

H. Kueck and N. de Freitas. Learning about individuals from group statistics. In *Conference on Uncertainty in Artificial Intelligence (UAI)*, 2005.

C. Leistner, A. Saffari, and H. Bischof. Miforests: Multiple-instance learning with randomized trees. In *European Conference on Computer Vision (ECCV)*, 2010.

F. Li and C. Sminchisescu. Convex multiple-instance learning by estimating likelihood ratio. *Advances in Neural Information Processing Systems (NIPS)*, pages 1360–1368, 2010.

W. Li, L. Duan, D. Xu, and I.W.H. Tsang. Text-based image retrieval using progressive multi-instance learning. In *International Conference on Computer Vision (ICCV)*, pages 2049–2055. IEEE, 2011.

J. Malik, S. Belongie, T. K. Leung, and J. Shi. Contour and texture analysis for image segmentation. *International Journal of Computer Vision*, 43(1):7–27, 2001.

O. Maron and T. Lozano-Pérez. A framework for multiple-instance learning. *Advances in Neural Information Processing Systems (NIPS)*, pages 570–576, 1998.

N. Quadrianto, A. Smola, T. Caetano, and Q. Le. Estimating labels from label proportions. *The Journal of Machine Learning Research*, 10:2349–2374, 2009.

S. Ray and M. Craven. Supervised versus multiple instance learning: An empirical comparison. In *Proceedings of the 22nd International Conference on Machine Learning (ICML)*, pages 697–704. ACM, 2005.

S. Rueping. Svm classifier estimation from group probabilities. In *Proc. of the 27th Int. Conf. on Machine Learning (ICML)*, 2010.

Daniel Tarlow, Kevin Swersky, Richard S Zemel, Ryan Prescott Adams, and Brendan J Frey. Fast exact inference for recursive cardinality models. In *Proceedings of the 28th Conference on Uncertainty in Artificial Intelligence (UAI)*, 2012.

A. Vedaldi and A. Zisserman. Efficient additive kernels via explicit feature maps. *IEEE Transactions on Pattern Analysis and Machine Intelligence*, 34(3):480–492, 2012.

P. Viola, J. Platt, and C. Zhang. Multiple instance boosting for object detection. In *NIPS*, 2006.

Jonathan Warrell and Philip HS Torr. Multiple-instance learning with structured bag models. In *Energy Minimization Methods in Computer Vision and Pattern Recognition*, pages 369–384. Springer, 2011.

Q. Zhang and S.A. Goldman. Em-dd: An improved multiple-instance learning technique. *Advances in Neural Information Processing Systems (NIPS)*, 14:1073–1080, 2002.

Z.H. Zhou, Y.Y. Sun, and Y.F. Li. Multi-instance learning by treating instances as non-iid samples. In *Proceedings of the 26th Annual International Conference on Machine Learning (ICML)*, pages 1249–1256. ACM, 2009.